\pdfoutput=1

\documentclass{article}

\usepackage{microtype}
\usepackage{graphicx}
\usepackage{booktabs} 
\usepackage{cite}
\usepackage{tikz}
\usetikzlibrary{calc,3d}
\usepackage{comment} 
\usepackage{amsfonts, amsmath, amstext, amssymb, mathrsfs, color, url}
\usepackage[colorlinks=false, linkcolor=blue]{hyperref}
\usepackage{array}
\usepackage{mdwmath}
\usepackage{mdwtab}
\usepackage{algpseudocode}
\usepackage{algorithm}
\usepackage{url}
\usepackage{setspace}
\usepackage{footnote}
\usepackage[skip=0pt]{caption}
\usepackage{subcaption}
\usepackage{float}
\usepackage{epstopdf}
\usepackage{multicol} 

\def\IB{{\mathbb B}}

\def\IN{{\mathbb N}}

\def\IP{{\mathbb P}}

\def\IR{{\mathbb R}}

\def\calA{{\mathcal A}}

\def\calC{{\mathcal C}}
\def\calD{{\mathcal D}}
\def\calE{{\mathcal E}}

\def\calI{{\mathcal I}}

\def\calK{{\mathcal K}}

\def\calM{{\mathcal M}}

\def\calO{{\mathcal O}}
\def\calP{{\mathcal P}}

\def\calU{{\mathcal U}}

\def\bA{\mbox {\boldmath $A$}}

\def\bI{\mbox {\boldmath $I$}}

\def\bL{\mbox {\boldmath $L$}}

\def\bQ{\mbox {\boldmath $Q$}}
\def\bR{\mbox {\boldmath $R$}}
\def\bS{\mbox {\boldmath $S$}}

\def\bX{\mbox {\boldmath $X$}}

\def\ba{\mbox {\boldmath $a$}}

\def\bc{\mbox {\boldmath $c$}}
\def\bd{\mbox {\boldmath $d$}}
\def\be{\mbox {\boldmath $e$}}

\def\bp{\mbox {\boldmath $p$}}

\def\br{\mbox {\boldmath $r$}}
\def\bs{\mbox {\boldmath $s$}}
\def\bt{\mbox {\boldmath $t$}}
\def\bu{\mbox {\boldmath $u$}}
\def\bv{\mbox {\boldmath $v$}}

\def\bx{\mbox {\boldmath $x$}}
\def\by{\mbox {\boldmath $y$}}
\def\bz{\mbox {\boldmath $z$}}

\def\bzer{{\pmb 0}}
\def\bone{{\pmb 1}}

\def\bpsi{\mbox{\boldmath $\psi$}}
\def\bthe{\mbox{\boldmath $\theta$}}

\newcommand{\upperRomannumeral}[1]{\uppercase\expandafter{\romannumeral#1}}

\newcommand{\st}{ \mbox{s.t.} }
\newcommand{\argmin}{ \textup{argmin} }

\newcommand{\rank}{ \textup{rank} }

\newcommand{\likes}{ \textup{likes} }
\newcommand{\dislikes}{ \textup{dislikes} }

\def\supp{{\textup{supp}}}
\def\tr{{\textup{tr}}}

\newcommand{\lrb}[1]{ \lbrace #1 \rbrace }

\def\({\left(}
\def\){\right)}

\newtheorem{lemma}{Lemma}

\newtheorem{proposition}{Proposition}

\newtheorem{corollary}{Corollary}

\newtheorem{problem}{Problem}

\makeatletter
\g@addto@macro\normalsize{%
  \setlength\abovedisplayskip{3pt}
  \setlength\belowdisplayskip{3pt}
  \setlength\abovedisplayshortskip{3pt}
  \setlength\belowdisplayshortskip{3pt}
}

\usepackage{hyperref}

\usepackage[accepted]{icml2018}

\icmltitlerunning{Learning Kolmogorov Models for Binary Random Variables }

\begin{document}

\twocolumn[
\icmltitle{Learning Kolmogorov Models for Binary Random Variables }



\icmlsetsymbol{equal}{*}

\begin{icmlauthorlist}
\icmlauthor{Hadi Ghauch}{to}
\icmlauthor{Mikael Skoglund}{to}
\icmlauthor{Hossein Shokri-Ghadikolaei}{to}
\icmlauthor{Carlo Fischione}{to}
\icmlauthor{Ali Sayed}{goo}
\end{icmlauthorlist}

\icmlaffiliation{to}{School of Electrical Engineering, Royal Institute of Technology, KTH, Stockholm}
\icmlaffiliation{goo}{School of Electrical, Ecole Polytechnique Federale de Lausanne }

\icmlcorrespondingauthor{Hadi Ghauch}{ghauch@kth.se}

\icmlkeywords{
  Kolmogorov Model; Learning; Interpretable Data models; Block Coordinate Descent; Semidefinite Relaxation
}

\vskip 0.3in
]



\printAffiliationsAndNotice{}  

\begin{abstract}
We summarize our recent findings~\citet{Ghauch_KER_17}, where  
we proposed a framework for learning a \emph{Kolmogorov model}, for a collection of binary random variables. 
More specifically, we derive conditions that \emph{link} outcomes of specific random variables, and extract valuable relations from the data. 
We also propose an algorithm for computing the model and show its \emph{first-order optimality}, despite the combinatorial nature of the learning problem.  
We apply the proposed algorithm to recommendation systems, although it is applicable to other scenarios. We believe that the work is a significant step toward interpretable machine~learning. 
\end{abstract}

\section{Introduction}
 {Machine learning} and {artificial intelligence} tools have permeated a large number of areas~\citep{Marr_AIApps_16}. 
 These tools are based on \emph{machine learning models}, which consist of learning an input-output mapping for a given dataset. 
Despite the plethora of models (e.g., matrix factorization~\citep{Koren_MF_09}, SVD-based models~\citep{Koren_SVD++_2008}, neural networks~\citep{LeCun_DL_15}, and models inspired from physics~\citep{Stark_QuantRec_16}), they lack \emph{interpretability}: not offering insight about the data, nor the underlying process. 

The work follows recent attempts at \emph{interpretable machine learning}~\citep{Kim_IntertableML_17}, where the lack of interpretablity may have serious consequences in {mission-critical systems}, ethics, and validation of computer-aided diagnosis~\citep{Kim_IntertableML_17}. 
While there is no consensus around the definition of interpretability, \emph{causality}~\citep{Lipton_ModelInter_16} is a vital component: it refers to {associations} within the data and information about the underlying data-generating process. We adopt the latter as our `definition' of \emph{interpretable model}, as one where data-to-data relations are \emph{accurately} discovered. 

We propose learning a so-called \emph{Kolmogorov Model (KM)} associated with a set of binary Random Variables (RVs). In addition to prediction, the interpretability of the model (as defined above) enables learning \emph{association rules}~\citep{Agrawal_MAR_1993}: We derive a sufficient conditions under which  the realization of one RV's outcome (deterministically) \emph{implies} the outcome of the other. In the context of recommendation systems, association rules identify groups of items, for which a user liking one item \emph{implies} that he/she likes {all other items} in the group. In cancer detection, the same rules identify groups of samples, for which the presence of DNA methylation in the group, implies its presence in all other samples. 
Additionally, these rules may provide insight into the physical  mechanisms underlying user preferences, and DNA methylation. 

We formulate the resulting problem as a \emph{coupled combinatorial problem}, decompose it into two subproblems using the \emph{Block-Coordinate Descent (BCD)} method, and we obtain provably optimal solutions for both. 
For the first one, we exploit the structure of linear programs on the unit simplex, to propose a low-complexity (yet optimal) \emph{Frank-Wolfe} algorithm~\citep{Frank_FW_56}. To bypass the inherent complexity of the second subproblem (combinatorial and NP-hard), we propose a semidefinite relaxation, and show its \emph{quasi-optimality} in recovering the optimal solution of the combinatorial subproblem. Finally, we show the convergence of our algorithm to a stationary point of the original problem. We refer the reader to~\citet{Ghauch_KER_17} for all the derivations/discussions. 

\section{System Model}
 \underline{Notation:} We use bold upper-case letters to denote matrices, bold lower-case letters to denote vectors, and calligraphic letters to denote sets. For a given matrix $\pmb{A}$, $[\bA]_{i,j}$ denotes element $(i,j)$,  $\tr(\bA)$ denotes its trace, $\Vert \pmb{A} \Vert_F$ its Frobenius norm, and $\bA^T$ its transpose. For a vector $\ba$, $[\ba]_i$  denotes element $i$, $[\ba]_{i:j}$ elements $i$ to $j$, and $\supp(\ba)$ its support. The inequality $\bx \leq \by$ holds element-wise. $\bI_n$ denotes the $n \times n$ identity matrix, $\pmb{1}$ and $\pmb{0}$ the all-one and all-zero vectors (of appropriate dimension). $\be_n$ is the $n$th elementary basis vector, 
$\calP = \lrb{ \bp \in \IR_{+}^{D} ~ | ~ \bone^T \bp = 1  } $ the unit probability simplex,
and $\lrb{n} = \lrb{1,\cdots,n}$.
\subsection{Problem Formulation } \label{sec:KM}
Consider a double-indexed set of binary \emph{Random Variables (RVs)}, $X_{u,i} \in \calA = \lrb{1,2} $, with indexes from $\calD = \lrb{ (u,i) ~|~ u \in \calU, i \in \calI}$. The RVs are defined on a sample space  $\Omega$, consisting of elementary events $\Omega= \lrb{ \omega_d ~|~ 1 \leq d \leq D } $. 
We denote by $\IP[ X_{u,i} = z ], ~ z \in \calA$, the probability that RV $X_{u,i}$ takes the value $z \in \calA$. Using that $\calA = \lrb{1,2}$, we write
\begin{align} \label{eq:KMbin}
 \IP[ X_{u,i} = 1 ] =  \bthe_u^T \bpsi_{i,1} \textup{ and }
 \IP[ X_{u,i} = 2 ] =  \bthe_u^T \bpsi_{i,2} ,  
 \end{align}
where $\bpsi_{i,1} + \bpsi_{i,2} = \bone$.  
$\bthe_u$ is a \emph{Probability Mass Function (PMF)} vector on the unit simplex, $\calP$, and $\lrb{ \bpsi_{i,1}, \bpsi_{i,2} } \in \IB^D $ are \emph{binary indicator vectors} representing the support of its probability measure. The model follows from established results in classical probability~\citep{Gray_proba_10}.
Since $X_{u,i}$ is binary, it is fully characterized by considering one outcome, 
\begin{align} \label{eq:KMbin1}
 &\IP[ X_{u,i} = 1 ] =  \bthe_u^T \bpsi_{i} , 
 \end{align}
\eqref{eq:KMbin} and \eqref{eq:KMbin1} are equivalent, and will be used interchangeably (dropping the $_z$ subscript of $\bpsi_i$ without any loss in generality). Thus, each RV $X_{u,i}$ is associated with (determined by) a \emph{PMF vector} $\bthe_u , u \in \calU$, and an \emph{indicator vector} $\bpsi_{i}, i \in \calI $. Notice that the model in~\eqref{eq:KMbin1} can approximate with arbitrarily small accuracy the measure corresponding to $\IP[]$, given a large enough $D$. 
\vspace{-.9cm}

\begin{problem}[Problem Statement]\rm
Let $p_{u,i}$ denote the empirical values of $\IP[ X_{u,i} = 1 ] $. 
We assume that $\lrb{p_{u,i}}$ are known for elements of a \emph{training set} $\calK \subseteq \calD $, where $\calK=\lrb{ (u,i) ~|~ (u,i) \in \calU \times \calI }$.
\footnote{Note that acquiring (estimates of) the empirical probabilities can be done via training, and the specific method is application-dependent (see Appendix~\ref{sec:app}).}
Given samples coming from the model in~\eqref{eq:KMbin1}, we wish to deduce the parameters of underlying probability distribution: find parameters of the KM, i.e.,  $ \lrb{\bpsi_{i}, \bthe_u  }$ that best describe  $\lrb{p_{u,i} ~|~ (u,i) \in \calK }$. The resulting problem is a \emph{fully parametric statistical inference} task. For tractability, we address it using the \emph{minimum mean-squared error} as a point estimator, which in turn results in minimizing $\sum_{(u,i) \in \calK}  (\IP[ X_{u,i} =1 ] - p_{u,i})^2  = \sum_{(u,i) \in \calK}  (  \bthe_u^T \bpsi_{i} - p_{u,i})^2$. Once computed, the optimal KM parameters can be used for prediction on a different set, and extracting statistical relations (among the RVs in $\calK$).  
The resulting optimization problem is 
\begin{align} \label{opt:PBvec}
(Q)
\begin{cases} 
\underset{ \lrb{\bpsi_i}, \lrb{\bthe_u} }{\min} \displaystyle\sum_{(u,i ) \in \calK}  \left(  \bthe_u^T \bpsi_{i} - p_{u,i} \right)^2 \triangleq \calE  \\
\st ~ \bthe_u \in \calP~, ~\bpsi_{i}  \in \IB^D , ~ \forall (u,i) \in \calK 
\end{cases} .
\end{align}
Our solution to this non-convex combinatorial problem is detailed in Section~\ref{sec:prop_alg}. 
\end{problem}


\subsection{Toy Example: Recommendation Systems}\label{ex:RecSys}
In this context, $X_{u,i}$ models the preference of user $u$ for item $i$, $(u,i) \in \calK$. 
Thus, $\IP[ X_{u,i} = 1 ]$ (resp. $\IP[ X_{u,i} = 2 ]$) models the probability that user $u$ likes (resp. dislikes) item $i$. Moreover, $\bthe_u$ determines the profile/taste of user $u$, $\bpsi_{i}$ is related to item $i$ (depending on genre, price, etc.), and the elementary events denote movie genres (e.g., $\omega_1=$ ``Action'', $ \omega_2 =$ ``SciFi'', etc.).
\footnote{The model is generic since interpreting the elementary events in context-dependent. For a coin toss, $\Omega = \lrb{\omega_1 , \omega_2 }$, the elementary events denote heads and tails, respectively.} 
Then, the corresponding empirical probability (i.e., training set) is obtained as $p_{u,i} \triangleq [\bR]_{u,i}/R_{\max} $ where $[\bR]_{u,i} \in \IN$ denotes the rating that user $u$ has provided for item $i$, and $R_{\max}$ the maximum  rating~\citep{Stark_NNMRec_15}. 

Consider a $10$-star ``recommendation system'', having $2$ users and $2$ items. 
We then find the $D$-dimensional ($D=3$) KM factorization to obtain , $\lrb{\bpsi_i}_{i=1}^2$ and $\lrb{\bthe_u}_{u=1}^2$.\footnote{$D$ is the size of the Kolmogorov space $\Omega$, the number of elementary events, and the dimension of the factorization (selected via cross-validation to minimize the test error).} 
To showcase the model's intuition, note that $p_{1,1}$, the probability that user $1$ likes movie $1$, is represented as  $\bpsi_{1}^T \bthe_1$. It is thus expressed as \emph{convex/stochastic} mixture of \emph{movie genres}, since elementary events are movie genres in this scenario. More generally, a KM represents a set of observed outcomes for RVs, as mixtures of elementary events. After finding the empirical probabilities from the rated entries (as above),  $(Q)$ is solved to learn  $\lrb{\bpsi_i}_{i=1}^2$ and $\lrb{\bthe_u}_{u=1}^2$, and an example result is shown below: 
\begin{align*} \small
\underbrace{
\begin{bmatrix}
0.3 & 0.5   \\
0.1 & 0.2
\end{bmatrix} }_{\lrb{p_{u,i}}}
= 
\begin{bmatrix}
\bthe_1^T \Big\lbrace 0.2   &  0.3    & 0.5 \\
\bthe_2^T \Big\lbrace 0.1    &  0.1    & 0.8 
\end{bmatrix}
\begin{bmatrix}
0 & 1 & \rbrace \textup{\small{Action}}   \\
1 & 1 & \rbrace \textup{\small{SciFi}}  \\
\underbrace{0}_{\bpsi_1} & \underbrace{0}_{\bpsi_2} & \rbrace \textup{\small{Drama}} 
\end{bmatrix}  \normalsize
\end{align*}

\subsection{Related Work} \label{sec:SOA}
Our proposal to model binary RVs as elementary events on a Kolmogorov space (Section~\ref{sec:KM}) is based on established results from classical probability theory. To our best knowledge, this specific formulation is novel. 
Because this model is rooted in probability theory, ~\eqref{eq:KMbin1} defines the outcome of a RV in the strict Kolmogorov sense (see definition in Section~\ref{sec:KM}), and the resulting association rules (Section~\ref{sec:ExpKM}) also hold analytically. This is the reason behind the versatility of the approach. We will also show that the combinatorial aspects of $(Q)$ are not a limitation.  

The inner product in~\eqref{eq:KMbin1} is reminiscent of factorization methods such as, \emph{Matrix Factorization (MF)}~\cite{Koren_MF_09}, \emph{Nonnegative Matrix Factorization (NMF)}~\cite{Lee_NMF_00}, SVD~\citep{Candes_SVT_10}, and physics-inspired techniques (e.g., Nonnegative Models (NNMs)~\citep{Stark_NNMRec_conf_16}). However, the inner product in these methods do not model RVs, 
in an analytical sense (Section~\ref{sec:KM}). Our method also generalizes K-means and some of its variants. Detailed discussions of the relation between our proposed method and these prior works is in Appendix~\ref{app:SoA}. 

\section{Proposed Algorithm} \label{sec:prop_alg}
We use the well-known Block-Coordinate Descent (BCD) framework to handle coupling in the cost function of $(Q)$ in~\eqref{opt:PBvec}. The method essentially splits $(Q)$ into two subproblems. We derive different methods for each subproblem, with provable accuracy, and show convergence of the algorithm. 
Given $\lrb{\bpsi_i^{(n)}}$ at iteration $n$, we first refine the current PMF estimate $\bthe_u$, as  
\begin{align*} 
(Q_1):\bthe_{u}^{(n+1)} \in 
\underset{\bthe_u \in \calP }{\argmin} ~f(\bthe_u) \triangleq\bthe_u^T \bQ_u^{(n)}  \bthe_u - 2 \bthe_u^T  \br_u^{(n)} ,
\end{align*}
where 
\begin{equation*} 
\resizebox{1\hsize}{!}{$  \bQ_u^{(n)} \triangleq \sum_{i \in \calI_K}  \bpsi_i^{(n)} \bpsi_i^{(n)^T},~  \br_u^{(n)} \triangleq \sum_{i \in \calI_K} \bpsi_i^{(n)}  p_{u,i} . ~~~~(e.1)$}
\end{equation*}
We then refine the current indicator vector estimate, $\bpsi_i$ as
\begin{align*} 
(Q_2):\lrb{\bpsi_{i}^{(n+1)}} \in \underset{\bpsi_i \in \IB^D}{\argmin} ~g(\bpsi_i) \triangleq  \bpsi_i^T \bS_i^{(n+1)} \bpsi_i - 2 \bpsi_i^T \bv_i^{(n+1)} 
\end{align*}
where 
\begin{equation*} 
\resizebox{1\hsize}{!}{$\bS_i^{(n+1)} \triangleq \sum_{u \in \calU_K}   \bthe_u^{(n+1)} \bthe_u^{(n+1)^T}, \bv_i^{(n+1)} \triangleq \sum_{u \in \calU_K} \bthe_u^{(n+1)} p_{u,i}~~(e.2)  $}
\end{equation*}

Moreover, $\calU_K$ and $\calI_K$ are defined as $\calK=\lrb{ (u,i) ~|~ u \in \calU_K \subset \calU ~, ~ i \in \calI_K \subset \calI }$. 
Next we describe the solution approach to each problem.

\subsection{Refine PMF Estimate} \label{sec:LCQP}
$(Q_{1})$ is a convex quadratic problem that can be solved by a variety of tools. However, we exploit its structure to greatly reduce the computational complexity: The Frank-Wolfe (FW) algorithm~\citep{Frank_FW_56} solves $(Q_1)$ as a succession of Linear Programs (LPs) over the unit simplex. While LP solvers generally have similar complexity as quadratic program solvers, solving an LP reduces to searching for the minimum index, when the LP is over the unit simplex. We formalize the algorithm, focusing on the original FW (detailed in~\citet{Jaggi_revisitingFW_13}[Algorithm 1]).  

While $\bthe_u$ should have two superscripts, $n$ for the BCD iteration and $k$ for the FW iteration, we only use $\bthe_u^{(k)} $. We first determine the \emph{descent direction}: 
\begin{align}
\bd_u^{(k)}  \in 
{\argmin}_{\bs} ~ \left( \nabla f(\bthe_u^{(k)}) \right)^T \bs  ~~\st ~~ \bs \in \calP ~.
\end{align}
The constraint $\bs \in \calP$ greatly simplifies the above LP, i.e.,  
\begin{align}
\bd_u^{(k)} = \be_{j^\star },~~ j^\star \in {\argmin}_{1 \leq j \leq D} ~  [ \nabla f(\bthe_u^{(k)})]_j ~.
\end{align}
The solution follows from LPs over the unit probability simplex (Proposition~\ref{prop:specLP}). Thus, finding the descent direction reduces to searching over the $D$-dimensional vector $\nabla f(\bthe_u^{(k)}) $ (done in~$\calO(D)$). 
Then, $\bthe_u^{(k)} $ is updated using a simple step size, $\alpha_u^{(k)} = k/(k+1) $ \citep{Jaggi_revisitingFW_13}. 
Table~\ref{alg:FWA} summarizes the FW procedure, and Proposition~\ref{prop:convFWA} shows its convergence, 
\begin{table}
\caption{Frank-Wolfe Procedure. } \label{alg:FWA} 
\begin{algorithmic}
\State \hrulefill
\Procedure{[$\bthe_u^\star$] = FWA }{ $\bQ_u ,  \br_u , \epsilon $ }  
\For{$k=1,2,..., I_{FW} $}
\State $\bd_u^{(k)} = \be_{j^\star }, ~\textup{where } j^\star = \underset{ 1 \leq j \leq D }{\argmin} ~  [\nabla f(\bthe_u^{(k)})]_j $
\State $\bthe_u^{(k+1)} = (1-\alpha_u^{(k)} ) \bthe_u^{(k)} + \alpha_u^{(k)} \bd_u^{(k)} $
\State Stop if $\Vert \bthe_u^{(k+1)} - \bthe_u^{(k)} \Vert  \leq \epsilon $
\EndFor
\EndProcedure
\State \hrulefill
\end{algorithmic}
\vspace{-.7cm}
\end{table}

\subsection{Refine Indicator Estimate } \label{sec:BinQP}

The NP-hard nature of $(Q_2)$ implies that relaxations are the only choice for a scalable solution. We thus propose a solution based on \emph{Semi-Definite Relaxation (SDR)} and randomization, and establish its \emph{quasi-optimality} for $(Q_2)$. 
We use the results of ~\citet{Ma_SDRML_02}[Sec IV-C]) and a series of reformulations to rewrite $(Q_2)$ in its equivalent form~\citep{Ghauch_KER_17}: 
\begin{align*}
\bX_i^\star \in 
\begin{cases} 
\underset{\bX_i}{\argmin} ~\tr(\tilde{\bS}_i \bX_i) \\
\st ~ \bX_i \succeq \bzer , ~ [\bX_i]_{k,k} = 1 , \forall k ~, ~~
\rank(\bX_i) = 1 
\end{cases}
\end{align*}
where $\bX_i = \bx_i \bx_i^T $,
$\tilde{\bS}_i = 
\begin{bmatrix}
(1/4)\bS_i  & -\tilde{\bt}_i/2 \\
-\tilde{\bt}_i^T/2 & 0
\end{bmatrix}
~,~ \bx_i =
\begin{bmatrix}
\bz_i \\
 w_i
\end{bmatrix}$
, $\bz_i = 2 \bpsi_i - \bone$,  $w_i \in \lrb{-1 , +1}$ is an auxiliary variable, and $\tilde{\bt}_i \triangleq  (\bv_i - (1/2)\bS_i \bone )$. 
The above problem is then relaxed into a convex SDP, 
\begin{align}\label{opt:Xsdr}
\bX_i^{\textup{(SDR)}} \in 
\begin{cases} 
\underset{\bX_i}{\argmin} ~\tr(\tilde{\bS}_i \bX_i)~. \\
\st ~ \bX_i \succeq \bzer , ~ [\bX_i]_{k,k} = 1 , \forall k ~  
\end{cases}
\end{align}
$\bX_i^{\textup{(SDR)}} $ may be solved using generic SDP solvers. 
Then, a randomization procedure~\citep{Ma_SDRML_02} extracts an approximate (binary) solution $\hat{\bpsi}_i$ of $(Q_2)$; see Table~\ref{alg:SDRR}.
\begin{table}
\vspace{-.5cm}
\caption{Semidefinite Relaxation + Randomization (SDR) Proc.} \label{alg:SDRR}
\begin{algorithmic}
\State \hrulefill
\Procedure{[$\hat{\bpsi}_i$] = SDR-wR }{ $\bS_i ,  \bt_i , M_{rnd} $  }  
\State // \emph{Repeat to approximate each} $\bpsi_i^\star,  \forall i \in \calI_K $
\State Solve~\eqref{opt:Xsdr} to find $\bX_i^{\textup{(SDR)}}$  
\State Factorize as $ \bX_i^{\textup{(SDR)}} = \bL_i^T \bL_i $ 
\For{$m=1,2,..., M_{rnd}$}
\State Generate zero-mean i.i.d Gaussian vector $\bu_i^{(m)}$ 
\State Compute $\hat{\bu}_i^{(m)} = \textup{sign}[ ~\bL_i^T \bu_i^{(m)} ~]  $
\EndFor 
\State Find $m^\star = \argmin_{ 1 \leq m \leq D+1 } ~~ \hat{\bu}_i^{(m)^T} \tilde{\bS}_i \hat{\bu}_i^{(m)} $ 
\State Compute $\hat{\bz_i} =  [\bu_i^{(m^\star)}]_{1:D} ~ [\bu_i^{(m^\star)}]_{D+1}   $
\State Approximate $\bpsi_i^\star $, as $\hat{\bpsi}_i = (\hat{\bz}_i + \bone)/2   $\
\EndProcedure
\State \hrulefill
\end{algorithmic}
\end{table}
This evidently raises the issue of the \emph{suboptimality gap} for SDR. We show in Proposition~\ref{prop:SDRopt} that SDR is optimal (asymptotically in $D$) in recovering the \emph{binary solution} of $(Q_2)$. Note that the performance bound in Proposition~\ref{prop:SDRopt} compares the quality of the approximate \emph{binary solution} offered by SDR, against the optimal solution of $(Q_2)$ (rather than just comparing the resulting cost functions).  

\subsection{Algorithm Description}
\begin{figure}
  \centering
  \includegraphics[ scale=.2]{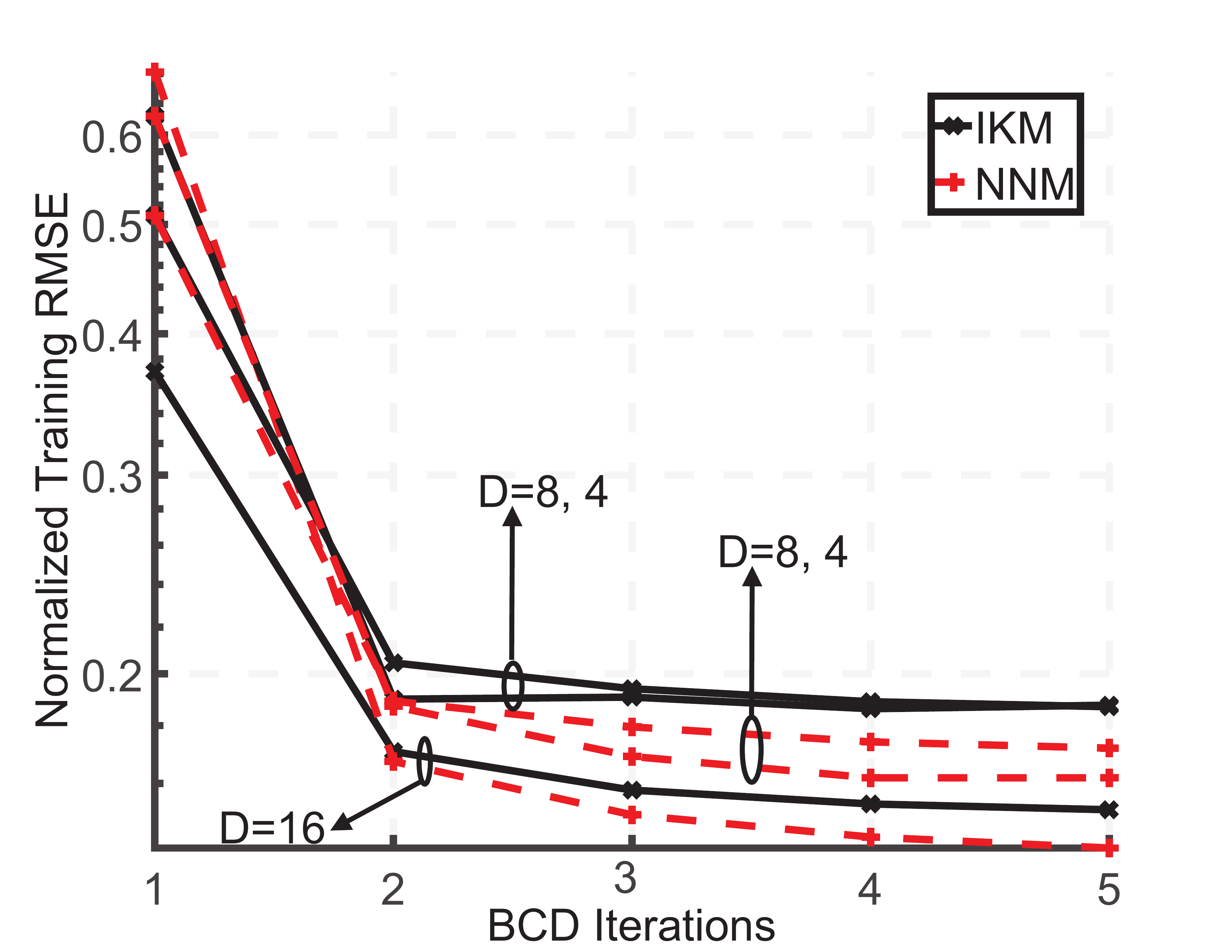}
  \caption{Training error for IKM for different values of $D$  (ML100K dataset) } 
  \label{fig:KM_vs_NNM}
\vspace{-.5cm}
\end{figure}

The BCD-based algorithm alternates between refining the indicator and PMF vectors (using the methods of Sec.~\ref{sec:BinQP} and Sec.~\ref{sec:LCQP}; see Algorithm~\ref{alg:IKM}. Its convergence to a stationary point of $(Q)$ is shown Lemma~\ref{lem:IKMconv}. 
Figure~\ref{fig:KM_vs_NNM} shows the convergence behavior of Algorithm~\ref{alg:IKM} for the 
ML100K dataset. The numerical setup and further results are provided in Appendix~\ref{app:num_res} (due to the lack of space). 

\begin{algorithm}
\caption{Iterative computation of KMs (IKM)} \label{alg:IKM}
\begin{algorithmic}
\State // \emph{Randomly Initialize} $\lrb{\bthe_u^{(1)} \in \calP }$   
\For{$n=1,2,... $} 
\State Compute $\bS_i^{(n)} $ and $ \bt_i^{(n)}$ from $(e.2)$
\State Update $\hat{\bpsi}_i^{(n)}$ = SDR-wR($\bS_i^{(n)} , \bt_i^{(n)} , M_{rnd} $ ), $\forall i \in \calI_K $ 
\State Compute $\bQ_u^{(n)} $ and $ \br_u^{(n)}$ from ~$(e.1)$
\State // \emph{Initialize FWA with $ \lrb{\bthe_u^{(n-1)}}$, from prev iter}
\State Update $\bthe_u^{(n)^\star}$ = FWA($\bQ_u^{(n)} ,  \br_u^{(n)} , \epsilon $ ), for all $u \in \calU_K $ 
\EndFor 
\end{algorithmic}
\end{algorithm}

\section{Interpretability via Association Rules}\label{sec:ExpKM}
Once a KM is found, we derive association rules that emerge from~\eqref{eq:KMbin} and~\eqref{eq:KMbin1}. 
\subsection{Association Rules}  
\begin{proposition}[Inclusion of Support Set] \label{prop:incset} \rm 
Consider two random variables $X_{u,i}$ and $X_{u,j}$  (belonging to the training set), whose KM are given by~\eqref{eq:KMbin}. 
If $\supp(\bpsi_{j}) \subseteq \supp(\bpsi_{i})$, then the following \emph{association rules} hold:
\begin{align} \label{eq:ARbin}
 &~ X_{u,i} = 1 \textup{~implies~}  X_{u,j} = 1  \\
 &~ X_{u,j} = 2 \textup{~implies~}  X_{u,i} = 2 \:.
\end{align}
\end{proposition}
For the toy example of Section~\ref{ex:RecSys}, note that $\supp(\bpsi_{1}) \subseteq \supp(\bpsi_2) $. Then, Proposition~\ref{prop:incset} yields: if user $1$ (or user $2$) likes movie $2$ implies he/she also likes movie $1$. 
Proposition~\ref{prop:incset} motivates us to look for cases where the support set condition trivially holds: when $\bpsi_i = \bone $, then $\supp(\bpsi_i) = \lrb{ D}$, and $\supp(\bpsi_j) \subseteq \supp(\bpsi_i)$ holds, for any choice of $\bpsi_j  \forall ~j \in \calI_K$, where $\calI_K$ defined in~Section~\ref{sec:prop_alg}. 
\begin{corollary}[Maximally Influential RVs] \rm \label{cor:MaxSupp}
Let $\lrb{ \bpsi_i,~ \bthe_u }_{(u,i) \in \calK}$ denote the KM associated with the outcome $1$ for $X_{u,i}$, i.e., $\lrb{X_{u,i} = 1}_{(u,i)\in \calK} $. 
We define $\calM = \lrb{ i ~ | ~ \bpsi_i = \bone  } $ as the set of RV outcomes with maximum support. Then, the condition $ \supp(\bpsi_j) \subseteq \supp(\bpsi_i)$ (Proposition~\ref{prop:incset}) holds trivially $ ~ \forall~j~\in~\calI_K$. It follows that the association rules in~\eqref{eq:ARbin} hold, for each $i~\in~\calM $.   
\end{corollary}
For maximally influential RVs, the realization of one outcome, $X_{u,i} = 1$, determines that of 
\emph{all RVs of the set} $\lrb{X_{u,j} = 1 ~|~ \forall j \in \calI_K }$. It is illustrated in Figure~\ref{fig:MIRV}. 
For a recommendation system, maximally influential RVs are items for which a user liking an item implies that he/she like all other items, in the training set. 
We underline that, unlike other probabilistic association rules, our approach provides deterministic rules. In Appendix~\ref{app:AdjMat}, we have provided efficient algorithmic approach to automatically mine these rules based on the adjacency matrix and influence score. Simulation results confirm the usefulness of our association rules ; see Appendix~\ref{app:num_res}. 

\section{Conclusion} 
We have proposed a framework for learning a Kolmogorov model, associated with a collection of binary RVs. Interpretability of the model (as defined by causality) was harnessed by deriving association rules, i.e., by finding sufficient conditions that bind outcomes of certain random variables. 
We also proposed an algorithm for computing a Kolmogorov model, a combinatorial non-convex problem, and showed its convergence to a stationary point of the problem, using block-coordinate descent. The combinatorial nature of the problem was addressed using a semi-definite relaxation, where we showed that it yields asymptomatically optimal solutions.  
Our results suggest that increased interpretability and improved prediction, do not cause a significant increase in complexity.

\appendix
\cleardoublepage

\setcounter{page}{1}
\setcounter{equation}{0}
\setcounter{figure}{0}
\renewcommand{\theequation}{A.\arabic{equation}}
\renewcommand\thefigure{A.\arabic{figure}}

\section{Supplementary Material for \\ Learning Kolmogorov Models for Binary Random Variables }

\subsection{Related Work} ~\label{app:SoA}
We position our work against other approaches (focusing on recommendation systems).

\textbf{Factorization Methods:}
Note that, $(Q)$ can be re-written as a low-rank matrix factorization problem, over the set of binary and stochastic matrices~\citep{Ghauch_KER_17}[Sec. 9.2]. Thus, the proposed approach is connected to factorization methods: 
\emph{Matrix Factorization (MF)}~\cite{Koren_MF_09}, \emph{Nonnegative Matrix Factorization (NMF)}~\cite{Lee_NMF_00}, SVD~\citep{Candes_SVT_10} (and their many variants/extensions) have gained {widespread applicability}, covering areas in sound processing, (medical) image reconstruction, recommendation systems and prediction problems~\cite{Davenport_MatRecSurvey_16}. These techniques assume that each element in $\calK$ is the inner product of two \emph{arbitrary vectors}. Thus, the model in~\eqref{eq:KMbin1} does not represent a RV (in a mathematical sense), when viewed in the context of the proposed model (Section~\ref{sec:KM}). Consequently, the analytical guarantees of Section~\ref{sec:ExpKM}, that yield the association rules, do not hold for general factorization methods: Though association rules may still be extracted, they are not as rooted in Kolmogorov probability theory, and lead to different statistical relations. Naturally, we wish the explore the association rules that arise from the proposed model. 

\textbf{Exact Factorization:}
Ideally, it is desirable to solve $(Q)$ exactly, i.e., find $\bthe_u, \bpsi_i $ satisfying $p_{u,i} = \bthe_u^T \bpsi_i, ~\forall (u,i) \in \calK$, over the training set $\calK$. 
While we are unaware of such results, we highlight a related variant where the factorization is solved exactly over the entire dataset $\calD$, i.e, $p_{u,i} = \bthe_u^T \bpsi_i, ~\forall (u,i) \in \calD$, using binary MF~\citep{Slawski_BMF_13}: It is not applicable when factorizing a subset of $\calD$, e.g., the training set $\calK$. Consequently, binary MF is unfit for prediction tasks.

\textbf{KMs as a generalization of K-Means:}  \label{opt:kmeans}
Consider a special case of $(Q)$, where $\bpsi_i$ is constrained to have one non-zero element. The resulting problem becomes the well-known \emph{$K$-means clustering}~\cite{Lloyd_LLS_82}. The K-means algorithms (and its variants K-medoids, fuzzy K-means and K-SVD), have become {pervasive in an abundance of applications} such as clustering, classification, image segmentation, DNA analysis, online dictionary learning, source coding, etc. Our approach \emph{generalizes K-means}, by allowing for overlapping clusters. While a  similar generalization of the classical K-means algorithm was considered in~\cite{Jiyoung_NEOKmeans_15}, the number of points per cluster is determined explicitly. In our approach however, the number of points per cluster is {optimized within the algorithm}.

\textbf{Nonnegative Models:} \label{sec:NNM}
\emph{Non-Negative Models (NNMs)}~\cite{Stark_NNMRec_conf_16}  are recent attempts at interpretable models. For reasons of computational tractability~\cite{Stark_NNMRec_conf_16}, NNMs are defined by relaxing~$\bpsi_i$ in~\eqref{eq:KMbin1}, to the unit hypercube.  
However, this relaxation \emph{impairs} the highly interpretable nature of the original model in~\eqref{eq:KMbin1}, making association rules \emph{less accurate}. 
Moreover, the relaxation implies that~\eqref{eq:KMbin1} no longer models the outcome of a random variable, thus \emph{limiting} its applicability to (many) problems where KMs are applicable. 

\subsection{Applications}\label{sec:app}
We briefly mention other applications. 

\textbf{Outage Prediction in Wireless Communication:}
Consider a network with several transmitters and receivers. In this setting, $X_{u,i}$ represents the state of the communication link, between transmitter $u$ and receiver $i$ (link $(u,i)$), and  $\IP[X_{u,i}=1]$ (resp.  $\IP[X_{u,i}=2]$) denotes the probability that it is ``good'' (resp. in outage). Then the corresponding KM, computed from $\calK$, can be used to predict the state of other links, in a different set. 
Moreover, the association rules in Section~\ref{sec:ExpKM} identify links in the network, where link $(u,i)$ good (resp. in outage) \emph{implies} that link $(u,j)$ good (resp. in outage). Thus the interpretability of KM provides valuable information on the network.

\textbf{DNA methylation for Cancer Detection:}
Recent investigations have suggested that DNA methylation, chemical changes in the DNA structure, may act as a cancer detection mechanism~\cite{Houseman_DNA_methylation_2012}. In this context, $p_{u,i}$ denotes the measured methylation level for location $i$ on the DNA, and sample $u$.
DNA methylation expresses $ p_{u,i} = \bpsi_i^T \bthe_u $, where $\bpsi_i$ is a binary vector indicating the presence or absence of DNA methylation at location $i$, and $\bthe_u$ is a PMF vector modeling the weight assigned to each location~\citep{Slawski_BMF_13}. From the perspective of KMs, $ \bpsi_i^T \bthe_u $ is the \emph{probability} that 
location $i$ and sample $u$ is methylated. 
Moreover, the association rules can identify groups of DNA locations for which the presence (resp. absence) of methylation in one location, implies its presence (or absence) for all other locations in the group. 
Note that, the above insights are {not possible} using conventional methylation analysis.

\subsection{Interpretable Aspects} \label{app:AdjMat}
Here we detail additional aspects of KMs, related to interpretability (via association rules). 

\begin{figure}
\centering
\includegraphics[trim={1cm 7cm 4cm 5cm}, clip, scale=.3]{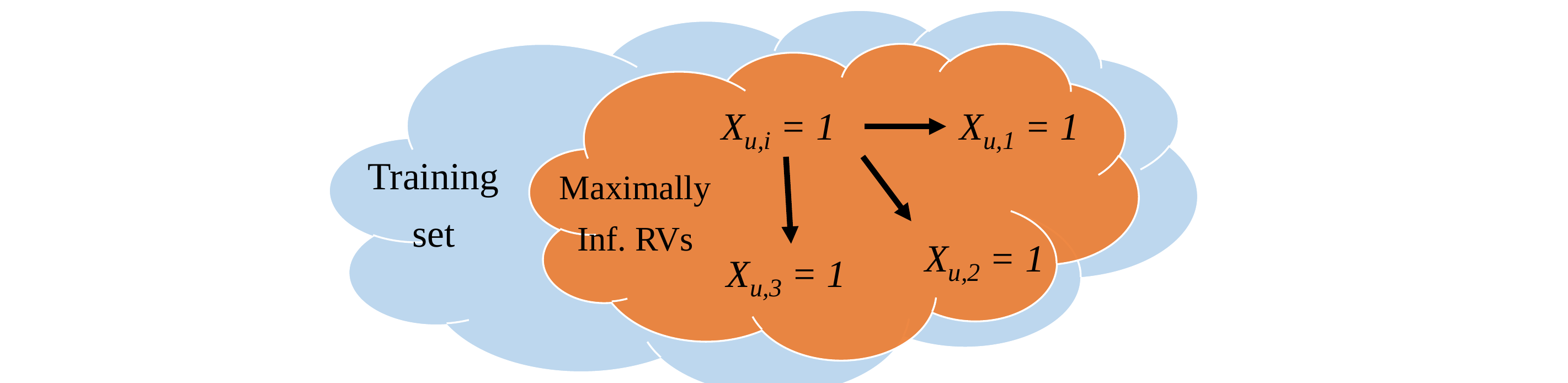}
  \caption{Association rules for maximally influential RVs } 
  \label{fig:MIRV}
\end{figure}

\textbf{Adjacency Matrix and Influence Score:} 
The association rules of Proposition~\ref{prop:incset} can be modeled using the so-called 
\emph{adjacency matrix}~$\bA \in \IB^{|\calI_K| \times |\calI_K|} $, define as 
\begin{align} \label{eq:AdjMat} 
 [\bA]_{i,j} = a_{i,j} = 
 \begin{cases}
 1, ~ \textup{ if } ~ \supp(\psi_j) \subseteq \supp(\psi_i)  \\
 0, ~ \textup{ otherwise }
 \end{cases} .
\end{align}
$[\bA]_{i,j} = 1$ denotes the inclusion of $X_{u,j}$ in $X_{u,i}$. 
In this case, the first outcome of $X_{u,i}$  implies the same outcome for $X_{u,j}$, and the second outcome for $X_{u,j}$ implies the second one for $X_{u,i}$ (as stated in Proposition ~\ref{prop:incset}), thereby implying \emph{coupling and mutual influence} among them (since $X_{u,i}$ influences $X_{u,j}$ and vice-versa). 
This raises the natural question of quantifying this coupling. We define an \emph{influence score} essentially counting (and normalizing) the number of pairs $X_{u,i}$ and $X_{u,j}$, satisfying the support set condition, 
\begin{align}
\beta_i = \frac{1}{\vert \calI_K \vert}  \sum_{ \substack{j \in \calI_K \\ j \neq i } } a_{i,j} \:.
\end{align}
Thus, we provide the following method for automatically mining these rules (presented in the context of recommendation systems).    
\begin{itemize} \setlength\itemsep{-.5em} 
\item Check the support set condition, via a pairwise search to check for pairs $\bpsi_i$ and $\bpsi_j$ satisfying  $\supp(\psi_j) \subseteq \supp(\psi_i) , ~\forall (i,j)~\in~\calI_K \times \calI_K , ~i~\neq~j$.
\item Build the adjacency matrix $\bA$, in~\eqref{eq:AdjMat} , and compute the influence score $\beta_i$
\item Find all pairs $(i,j)$ such that $a_{i,j} = 1 $: for each of these pairs the following holds (from Proposition~\ref{prop:incset}), 
\begin{align}  \label{eq:Aij}
\begin{cases}
  ~ [u ~\likes ~i ] &\textup{~implies~}  [u ~\likes ~j]  \\
  [u ~\dislikes ~j ] & \textup{~implies~} [u ~\dislikes ~i]  
 \end{cases}
\end{align}
\item Identify, if possible, maximally influential RVs (Corollary~\ref{cor:MaxSupp}), having the all-one indicator vector, i.e., $\calM= \lrb{ i ~ \vert ~ \bpsi_i = \bone }$. 
For each of them, the relations in~\eqref{eq:Aij} hold for \emph{all other items in the collection}
\end{itemize}

\textbf{Practical Issues Regarding Interpretability} 
 We recall that the proposed SDR method was shown to be quasi-optimal in providing approximate \emph{binary solutions} to $(Q_2)$. Thus, the relaxation does not affect the interpretability, in the sense that Proposition~\ref{prop:incset} and Corollary~\ref{cor:MaxSupp} still hold.
However, another remark is in order.  While the derivations pertaining to association rules (Section~\ref{sec:ExpKM}) assume globally optimal solutions to $(Q)$ - an NP-hard problem, Algorithm~\ref{alg:IKM} guarantees locally optimal ones. Thus, a bound on the gap between these solutions is needed. We highlight this issue as an interesting topic for further investigation.

\subsection{Variations and Special Cases}
\textbf{Learning RVs with Common Support:}
We underline some interesting special case of the proposed approach, namely, when all the RVs have the same   support, i.e., $\bpsi_1 = \cdots = \bpsi_D \triangleq \bpsi$.  This reduces to learning KMs, for RVs having common support. 

\textbf{Learning a sequence of RVs:}
Consider a special case of Sec.~\ref{sec:KM}, where we learn a KM for a \emph{sequence} of binary RVs, $\lrb{X_u ~|~  u \in \calU} $, from observing samples from the training set, $\lrb{ p_u  ~|~  u \in \calU_K} $. 
The KM in~\eqref{eq:KMbin1} reduces to $\IP[ X_{u} = 1 ] =  \bthe_u^T \bpsi$, and resulting optimization becomes: 
\begin{align} 
\begin{cases} 
\underset{\lrb{\bthe_u} , \bpsi }{\min} \sum_{ u \in \calU_K} (  \bthe_u^T \bpsi - p_{u} )^2  \\
\st ~ \bthe_u \in \calP~, ~\forall u \in \calU_K , ~\bpsi  \in \IB^D 
\end{cases}
\end{align}
The BCD-based solution approach is still applicable in this case, though many simplifications are possible. 

\subsection{Practical Aspects} \label{sec:complexity}
\textbf{Including Regularization Parameters:}
Regularization parameters for $\bthe_u$ and $\bpsi_i$, are needed for prediction to avoid over-fitting~\citep{Bishop_PatternRec_06}[Sec. 1.1]. They can be included without any changes to the solution method.  
An $\ell_2-$regularization can be included in $(Q_1)$:
\begin{align} \label{eq:freg}
f(\bthe_u) 
&= \bthe_u^T( \bQ_u + \lambda_u \bI_D  ) \bthe_u - 2 \bthe_u^T  \br_u  +  \gamma_{u} ~, 
\end{align}
where the regularizer $\lambda_u \geq 0$ is absorbed into a ``new'' matrix $( \bQ_u + \lambda_u \bI_D  ) $. While desirable, an $\ell_1-$regularization for $\bthe_u$ would not work, since $\bthe_u \in \calP$. 
Similarly, an $\ell_1$-regularization for $(Q_{2})$ is,  
\begin{align}
g(\bpsi_i)  
&=\bpsi_i^T \bS_i \bpsi_i - 2 ( {\bv_i -  (\mu_i/2) \bone} )^T \bpsi_i  +  \gamma_{i} ~,
\end{align}
where the regularizer $\mu_i$ is absorbed into the linear term, since $\mu_i \Vert  \bpsi_i \Vert_1 = \mu_i  \bone^T  \bpsi_i  $, for $\bpsi_i$ binary. 
 
\textbf{Computational Complexity:} 
The computational complexity of Algorithm~\ref{alg:IKM} is dominated by the SDP solution in~\eqref{opt:Xsdr},  
$\approx \calO(D^{4.5})$ for medium accuracy solutions (keeping in mind the negligible cost of the FW method). 
Thus, the total cost (per iteration) of Algorithm~\ref{alg:IKM} is  $\calC_{\textup{KM}}  \approx \calO(D^{4.5}) $. The added complexity compared to MF, NMF, NNM and SVD++ ($\approx \calO(D^{3})$) is not significant, keeping in mind that  $D \ll \min(|\calI_K| ~, |\calU_K| )$. Moreover, complexity reduction techniques (for the SDP solution) can be investigated. Finally, proposed method yields problems that decouple , thereby significantly speeding up the computation due to \emph{parallelization}.   

\textbf{Non-stationary distributions:} 
The proposed method assumes that distributions of the RVs (in the training set) are stationary: Indeed, scenarios with \emph{time-varying distributions} are a limitation (and interesting future directions). However, in learning it is quite common to assume that the data-generating distribution is stationary.

\subsection{Main Results} \label{app:mainres}
Below, we summarized the results used in the paper; see~\citet{Ghauch_KER_17} for the proofs. 

\textbf{LPs over the Unit Probability Simplex:}
We use following known result to find the descent direction for the FW method (the proof is known).
 \begin{proposition} \label{prop:specLP} 
 Consider the following Linear Program (LP), 
 \begin{align*}  
   (P_{PS}) ~~ \bx^\star = \underset{\bx \in \IR^{n} }{\argmin} \ \bc^T \bx  , \ \st \   \bone^T \bx =1 , \ \bx \geq \bzer  \quad
 \end{align*}
  Its optimal solution is given by 
 \begin{align*}  
 \bx^\star = \be_{j^\star} ~, \textup{where } ~j^\star = \ \argmin_{ 1 \leq j \leq n } \ \bc^T \be_j  
 \end{align*}
 Thus, the solution reduces to searching over the vector $\bc$.\hfill $\Box$
\end{proposition}

\textbf{Convergence of FW algorithm:}
We show the convergence of the FW algorithm (Table~\ref{alg:FWA}).  
\begin{proposition}  \label{prop:convFWA}
Let $\bthe_u^{\star}$ be the optimal solution to $(Q_{1})$. Then the sequence of iterates $\lrb{\bthe_u^{(k)} } $ satisfies ~\citep{Jaggi_revisitingFW_13}[Theorem 1], 
\begin{align*}
\Vert f(\bthe_u^{(k+1)}) - f(\bthe_u^{\star}) \Vert_2 \leq \calO(1/k), ~k = 1, 2, \cdots \hfill \Box
\end{align*}
\end{proposition}
\underline{Proof:} The linear convergence rate for all FW variants, was proved in~\citet{Jaggi_revisitingFW_13}[Theorem 1]. 

\textbf{Quasi-optimality of SDR:}
The question was studied extensively in the context of binary detection for multi-antenna communication~\citep{Rasmussen_SDR_CDMA_01}. Interestingly, $(Q_2)$ can be recast as a {noiseless} binary detection problem, where SDR has been to be optimal. The results is formalized below.  
\begin{proposition}  \label{prop:SDRopt}
Let $g(\bpsi_i^\star )$ and $g(\hat{\bpsi}_i) $ denote the optimal solutions to the binary QP in $(Q_{2})$, and its SDR after randomization (Table~\ref{alg:SDRR}), respectively. The approximation quality is defined as~\citep{Luo_SDR_2010},  
\begin{align}
\eta \leq ~{g(\bpsi_i^\star )} / {g(\hat{\bpsi}_i) } ~\leq 1 . 
\end{align}
It holds that $\eta=1$, with probability $1- \exp^{-\calO(D)} $, asymptotically in $D$. Thus, the relaxation is quasi-optimal.\hfill $\Box$  
\end{proposition}
\underline{Proof:} See~\citep{Ghauch_KER_17}.

\textbf{Convergence of IKM:}
\begin{lemma} \label{lem:IKMconv}
Let $t_n \triangleq \calE ( \lrb{\bpsi_{i}^{(n)}} ,  \lrb{\bthe_{u}^{(n)}}  ) , \ n = 1,2,... $ be the sequence of iterates, resulting from the updates in IKM. Then, $\lrb{t_n}$ is non-increasing in $n$, and converges to a stationary point of $(Q)$ in~\eqref{opt:PBvec}, almost surely. \hfill $\Box$
\end{lemma}
\underline{Proof:} The convergence is shown in~\citep{Ghauch_KER_17}.

\subsection{Numerical Results} \label{app:num_res}


\textbf{Experimental Setup:}
The \emph{training set} $\calK$, is chosen as the MovieLens 100K (ML100K), with $U=943$ users and $I=1682$ items, split into $80\%$ for training and $20\%$ for testing. 
Let $\lrb{\hat{\bpsi}_i}, \lrb{ \hat{\bthe}_u}  $  the output of Algorithm~\ref{alg:IKM}, after $5$ iterations (used to predict $p_{u,i}$ over the test set). 
For benchmarking, we factorize the rating matrix using MF~\citep{Koren_MF_09}, NMF~\citep{Lee_NMF_00}, SVD++\citep{Koren_SVD++_2008} (ensuring the dimension of the factorization, $k$, is close to $D$). The implementation and results use the MyMediaLite package~\citep{Gantner_MyMediaLite_11}, and the corresponding performance results~ are available \url{http://www.mymedialite.net/examples/datasets.html}. We also benchmark against the NNM algorithm in~\citet{Stark_NNMRec_15}, and the classical $K$-means (K-M) algorithm. 

\textbf{Training Performance:}
We first evaluate the performance of Algorithm~\ref{alg:IKM} on artificial training data, i.e., $p_{u,i} \in \calK = \lrb{U=20} \times \lrb{I=40} $ where $\lrb{p_{u,i}}$ are i.i.d. and uniformly chosen on the unit interval. 
We benchmark against a variant on Algorithm~\ref{alg:IKM}, where the SDR solution for $Q_2$ is replaced by an \emph{exhaustive search}. As the data is artificial, the resulting matrix does not have any missing entries: we also include the \emph{binary matrix factorization (BMF)} in~\citet{Slawski_BMF_13}[Algorithm 2].
\begin{table} [h]
\caption{Error rate for SDR ($U=20, I=40$). } \label{tab:SDRerr}
\small
\begin{center}
\begin{tabular}{ |c|c|c|c|}
 \hline
     &  $ D=4 $                        & $D=8  $                      & $ D=10 $                  
\cr \hline        
  SDR Accur. $\times 10^{-3}$  &  $ 7.5 $                        & $4.4  $                      & $ 4.0 $            
  \cr \hline
\end{tabular} 
\end{center}
\normalsize
\end{table}
Table~\eqref{tab:SDRerr} is a numerical validation of Proposition~\ref{prop:SDRopt} where we computed the error rate of SDR (compared to the exhaustive search), aggregated over all iterations. We observe that the approximation error decreases, with increasing $D$ (following Proposition~\ref{prop:SDRopt}). 
 
\begin{figure}
  \centering
  \includegraphics[trim={1cm 0 0 2,5cm}, clip, scale=.2]{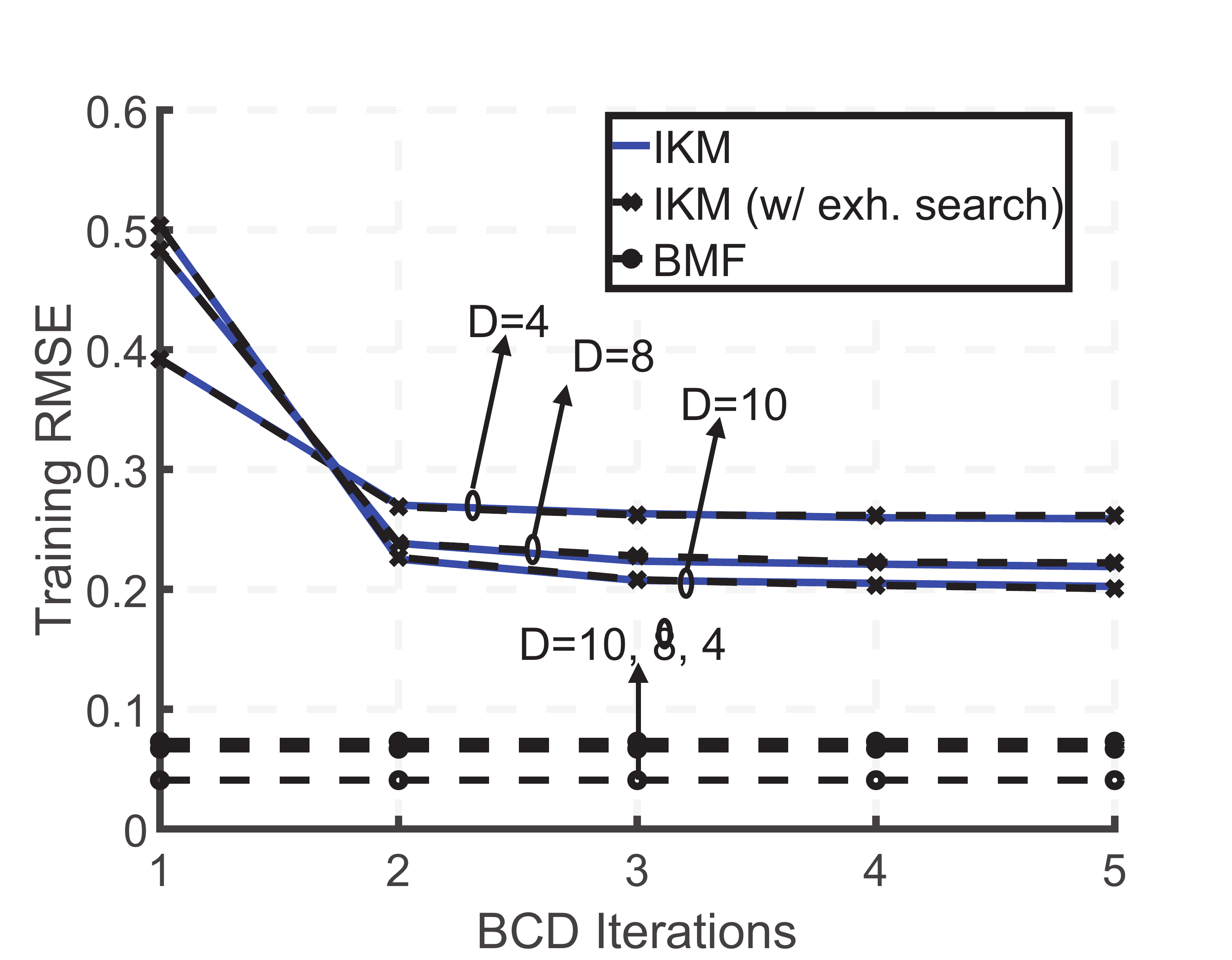}
  \caption{Training error vs number of iterations ($U=20, I=40$) } 
  \label{fig:SDRerr}
\end{figure}

Following the same setup and benchmarks, Fig~\ref{fig:SDRerr} shows the resulting normalized training RMSE training error, $\textup{RMSE} = ( \sum_{(u,i) \in \calK } \vert p_{u,i}- \hat{\bthe}_u^T \hat{\bpsi}_i \vert^2/|{\calK}| )^{1/2}$, for several values of $D$. We observe that the monotone convergence in Lemma~\ref{lem:IKMconv} is validated numerically, and that the training error decays with increasing model size, $D$. While the performance of IKM (first-order optimality guarantee) is indistinguishable from its exhaustive search variant, there is large gap compared to BMF (globally optimal). Unfortunately, BMF does not work with missing data, and is inapplicable to prediction \citep{Slawski_BMF_13}. The same conclusions hold when testing Algorithm~\ref{alg:IKM} on the ML100K (Fig.~\ref{fig:KM_vs_NNM}).

\begin{table} 
\tiny
\caption{Normalized RMSE values for test set (ML100K). The dimension of factorization for MF/SVD++, $k$, is equal to $D$ (unless stated in the corresponding entry).   } \label{tab:RMSEres}
\begin{center}
\begin{tabular}{ |c|c|c|c|c|}
 \hline
       &  $ D=4 $                        & $D=8  $                      & $ D=16 $                     & $D=24$  \\
 \hline        
\textbf{KM}  &  $ {0.199}$       &  $\textbf{0.2013} $    & $\textbf{0.1900}$ & $\textbf{0.1861}$ \\ 
NNM  & $ \textbf{0.194}$ &  $0.2255$    & $0.2057$                & $0.2118$ \\ 
MF  & $ 0.229$ &  $0.228(k=10) $    & $-$                & $0.226(k=40)$ \\ 
SVD++  & $ 0.228$       &  $0.227(k=10)$    & $0.227(k=20)$ & $0.226(k=50)$ \\ 
K-M  & $ 0.210$ &  $.2096 $    & $0.2105$ & $0.2105$ \\ 
NMF  &  $ -$       &  $- $    & $-$ & $0.192(k=100)$ \\ 
\hline
\end{tabular} 
\end{center}
\normalsize
\end{table}

\begin{figure}
\centering
  \includegraphics[trim={1cm 0 0 2cm}, clip, scale=.2]{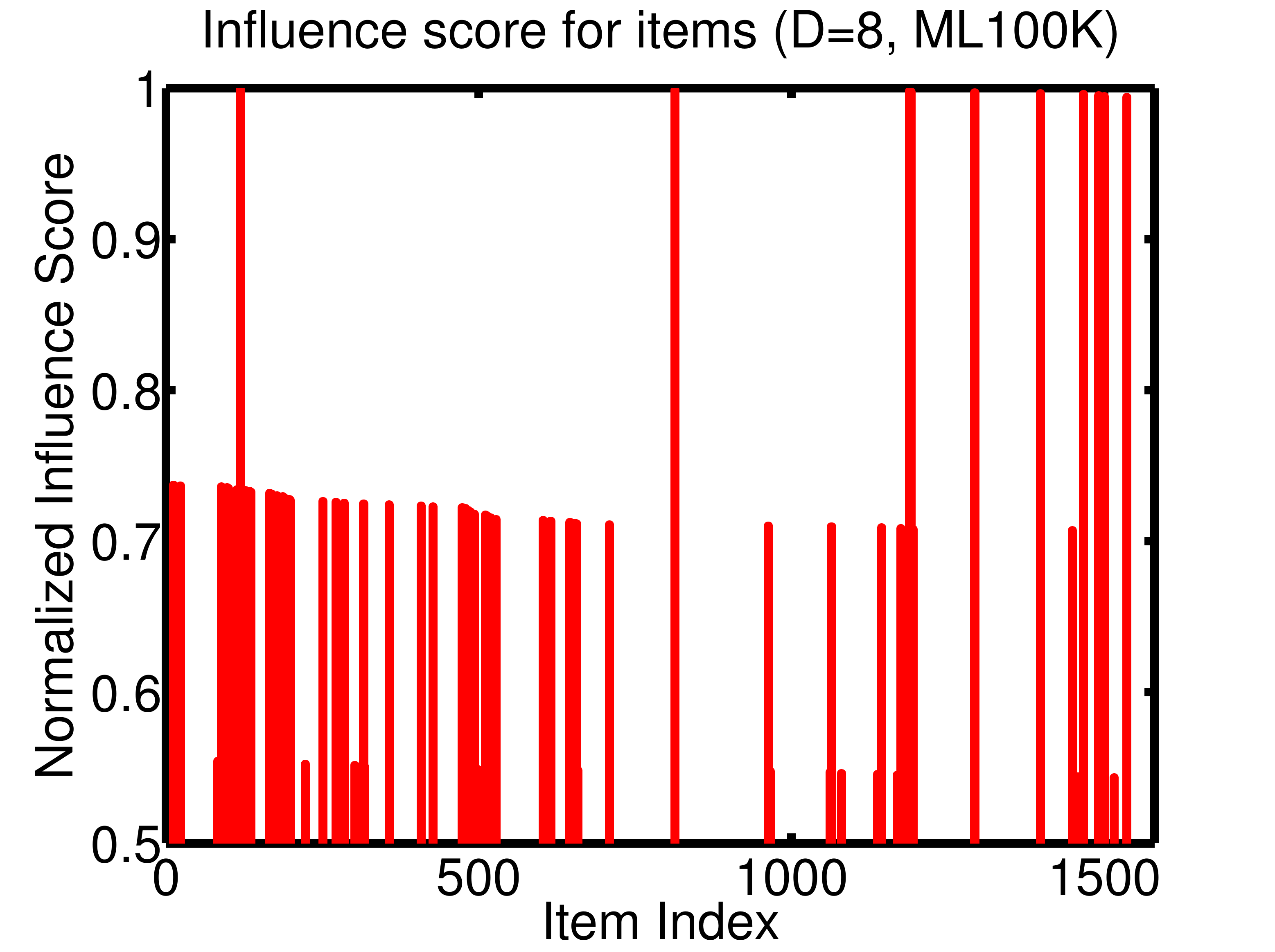}
  \caption{Influence score for items having $\beta_i \geq 0.5$ ($D=8$, ML100K dataset) } 
  \label{fig:infscore}
\end{figure}

\textbf{Interpretability of KMs:}
We numerically evaluate the method for finding association rules (Section~\ref{app:AdjMat}), on the ML100K dataset, with the resulting influence scores shown in Fig.~\ref{fig:infscore}. 
We first identify the set of maximally influential items, 
$\calM=\lrb{ 119  ,       814  ,      1188   ,     1190   ,     1290  ,      1393  ,      1462  ,      1486    ,    1494   ,   $ $  1530 ,1590    ,    1638 }$.  
For each of these items, a user liking one given item, implies he/she likes all other items in the training set. 
Interestingly, these results remain the same when $D=24$, thereby suggesting that procedure for mining association rules is quite stable.

\textbf{Prediction Performance:} 
Since the range of the predicted variable is different for MF/NMF/SVD++, and KM/NNM, we use the normalized test   RMSE, i.e., $\textup{NRMSE} = \eta 
( \sum_{(u,i) \in \bar{\calK} } \vert [\bR]_{(u,i)} -\hat{R}_{u,i} \vert^2/|\bar{\calK}|)^{1/2} $  
where $\bar{\calK}$ is the test set, and $\eta = (R_{\max} - R_{\min})^{-1}=1/4 $ is the normalization for MF/NMF/SVD++. 
For KMs/NNMs the same metric reduces to 
$\textup{NRMSE} =  \left( \sum_{(u,i) \in \bar{\calK} } \vert [\bR]_{(u,i)}/R_{\max} - \hat{\bthe}_u^T \hat{\bpsi}_i \vert^2/|\bar{\calK}| \right)^{1/2} $.
The best values for $\lambda_u$ and $\mu_i$, were picked from a coarse two-dimensional grid by cross-validation, using a held-out validation set. The Normalized RMSE results are shown in Table~\ref{tab:RMSEres}. 
We observe a significant gap between KMs, and well known collaborative filtering methods, especially as $D$ increases. Moreover, the drop in performance for NNMs for increasing $D$ may be due to over-fitting. 


\bibliography{ref_hadi_merged}
\bibliographystyle{icml2018}

\end{document}